\documentclass[11pt]{article}   
\usepackage{graphicx,colacl}  

\title{Integrating Text Plans for Conciseness and Coherence\thanks{This work was supported by the National Library of
Medicine under grant R01-LM-05764-01. We thank Bonnie Webber and
John Clarke for their suggestions and advice during the course of this
research.}}

\author{Terrence Harvey \and Sandra Carberry\\Department of Computer
Science\\University of Delaware\\Newark, DE 19716\\ \{harvey,carberry\}@cis.udel.edu}

\begin{document}
\maketitle
\bibliographystyle{acl}
\begin{abstract}Our experience with a critiquing system shows that when the
system detects problems with the user's performance, multiple
critiques are often produced.  Analysis of a corpus of actual
critiques revealed that even though each individual critique is
concise and coherent, the set of critiques as a whole may exhibit
several problems that detract from conciseness and coherence, and
consequently assimilation.  Thus a text planner was needed that could
integrate the text plans for individual communicative goals to produce
an overall text plan representing a concise, coherent message. 

This paper presents our general rule-based system for accomplishing
this task.  The system takes as input a \emph{set} of individual text
plans represented as RST-style trees, and produces a smaller set of
more complex trees representing integrated messages that still achieve
the multiple communicative goals of the individual text plans.
Domain-independent rules are used to capture strategies across
domains, while the facility for addition of domain-dependent rules
enables the system to be tuned to the requirements of a particular
domain.  The system has been tested on a corpus of critiques in the
domain of trauma care.
\end{abstract}

\section{Overview}
Many natural language systems have been developed to
generate coherent text plans
\cite{Moore93b,Hovy91a,Wanner96,Zukerman95a}. However, none has the
ability to take a set of independently generated yet inter-related
text plans and produce integrated plans that realize all of the
communicative goals in a concise and coherent manner.

 
{\em RTPI} (Rule-based Text Plan Integrator) was designed to perform
this task. The need for coherence requires that the system be able to
identify and resolve conflict across multiple, independent text plans,
and exploit relations between communicative goals. Conciseness
requires the ability to aggregate and subsume communicative
goals. Although our work was motivated by the need to produce
coherent, integrated messages from the individual critiques produced
by a decision support system for emergency center trauma care, this
same task will arise in future systems as they make use of independent
modules that need to communicate with a user. Thus the system should
have simple, domain-independent rules, but should also be flexible
enough to allow the addition of rules specific to the domain at hand.

This paper describes {\em RTPI} and our initial implementation that
works with the kinds of text plans representative of a critiquing
system. While our examples are taken from the domain of trauma care,
the domain-independent rules make the system applicable to other
domains of critiquing and instruction as well. The motivation behind
{\em RTPI} is presented in Section~\ref{sec-mot}, and Section
~\ref{sec-other} contrasts it with other work. Then we describe the
system's parameters that allow flexible response in multiple
environments (Section~\ref{sec-params}). The heart of the system is
{\em RTPI}'s domain-independent rule base (Section~\ref{sec-rules})
for integrating text plans. The implemented algorithm and the results
of its application are presented last.

\vspace{-.05in}
\section{Motivation}
\label{sec-mot}
\vspace{-.05in}

TraumAID \cite{Webber92} is a decision support system for addressing
the initial definitive management of multiple trauma. TraumaTIQ
\cite{Gertner96} is a module that infers a physician's plan for
managing patient care, compares it to TraumAID's plan, and critiques
significant differences between them. TraumaTIQ recognizes four
classes of differences: errors of omission, errors of commission,
scheduling errors, and procedure choice errors. Experimentation with
TraumaTIQ showed that when the physician's plan is deficient, several
problems are generally detected, and thus multiple critiques are
independently produced.

We analyzed 5361 individual critiques comprising 753 critique sets
produced by TraumaTIQ on actual cases of trauma care. A critique set
represents the critiques that are produced at a particular point in a
case. While each critique was coherent and concise in isolation, we
found several problems within critique sets: some critiques detracted from
others in the critique set; some would make more sense if they took
explicit account of other critiques appearing earlier in the set; and
there was informational overlap among critiques. 

Our analysis revealed 22 common patterns of inter-related critiques,
each pattern covering some subset of a critique set. While we
initially developed a domain-dependent system, TraumaGEN, that
operated directly on the logical form of the critiques produced by
TraumaTIQ, we noted that many of the patterns were more
generally applicable, and that the problems we were addressing would
also arise in other sophisticated systems that distribute their
processing across multiple independent modules, each of which may need
to communicate with the user. While such systems could be designed to
try to prevent problems of this kind from arising, the result would be
less modular, more complex, and more difficult to extend.

Thus we developed {\em RTPI}, a system for constructing a set of
integrated RST-style text plans from a set of individual text
plans. {\em RTPI} contains a set of domain-independent rules, along
with adjustable parameters that determine when and how rules are
invoked. In addition, {\em RTPI} allows the addition of
domain-dependent rules, so the system can account for interactions and
strategies particular to a domain.

\vspace{-.05in}
\section{Other Work}
\label{sec-other}
\vspace{-.05in}

\setlength{\textfloatsep}{0.05in}
\begin{figure}
\begin{center}
\begin{tabular} {p{3.0in}}
\underline{TraumaTIQ critiques:}\\
{\em Caution: check for medication allergies and do a laparotomy immediately to treat the intra-abdominal injury.}
\\
{\em Consider checking for medication allergies now to treat a
possible GI tract injury.} 
\\
{\em Please remember to check for medication allergies before you give
antibiotics. } 
\\
\\
\underline{Message from {\em RTPI} integrated plan:}\\
{\em Caution: check for medication allergies to treat the
intra-abdominal injury and a possible GI tract injury, and do it
before giving antibiotics. Then do a laparotomy to complete treating
the intra-abdominal injury. } \\ 
\end{tabular}
\end{center}
\caption{Result of communicative goal aggregation.}
\rule{3.1in}{.010in}
\label{agg-text}
\end{figure}
\vspace{-.05in}

The idea of domain-independent text planning rules is not new.
\newcite{appelt85a} used ``interactions typical of linguistic
actions'' to design critics for action subsumption in KAMP. REVISOR
\cite{callaway97} used domain-independent operators for revision of a
text plan for explanation. Because our rules operate on full RST-style
text plans that include communicative goals, the rules can be designed
to integrate the text plans in ways that still satisfy those goals.

The Sentence Planner \cite{Wanner96} uses rules to refine a single
initial tree representation. In contrast, {\em RTPI} operates on
\emph{sets} of complete, independent text plan trees. And while
REVISOR handles clause aggregation, and Sentence Planner removes
redundancies by aggregating neighboring expressions, neither of them
addresses the aggregation of communicative goals (often requiring
reorganization), the revision and integration of text plans to remove
conflict, or the exploiting of relations between communicative goals
as done by {\em RTPI}. Similarly, WISHFUL \cite{Zukerman95a} includes
an optimization phase during which it chooses the optimal way to
achieve a set of related communicative goals. However, the system can
choose to eliminate propositions and does not have to deal with
potential conflict within the information to be conveyed.

\vspace{-.05in}
\section{System Parameters}
\label{sec-params}
\vspace{-.05in}

Although {\em RTPI}'s rules are intended to be domain-independent,
environmental factors such as the purpose of the messages and the
social role of the system affect how individual text plans should be
integrated. For example, if the system's purpose is to provide
directions for performing a task, then an ordered set of actions will
be acceptable; in contrast, if the system's purpose is decision
support, with the user retaining responsibility for the selected
actions, then a better organization will be one in which actions are
grouped in terms of the objectives they achieve (see
Section~\ref{sec-agg}). Similarly, in some environments it might be
reasonable to resolve conflict by omitting communicative goals that
conflict with the system's action recommendations, while in other
environments such omission is undesirable (see
Section~\ref{sec-conflict}).

{\em RTPI} has a set of system parameters that capture these
environmental factors. These parameters affect what rules are
applied, and in some cases how they are applied. They
allow characteristics of the output text plans to be tailored to broad
classes of domains, giving the system the flexibility to be effective
over a wide range of problems.

\vspace{-.05in}
\section{The Rule-Base}
\label{sec-rules}
\vspace{-.05in} 

{\em RTPI}'s input consists of a set of text plans, each of which has
a top-level communicative goal. Rhetorical Structure Theory
\cite{mannthompson87} posits that a coherent text plan consists of
segments related to one another by rhetorical relations such as
\textsc{motivation} or \textsc{background}. Each text plan presented
to {\em RTPI} is a tree structure in which individual nodes are related by
RST-style relations. The top-level communicative goal for each text
plan is expressed as an intended effect on the user's mental state
\cite{Moore95}, such as 
\scriptsize\verb+(GOAL USER (DO ACTION27))+\normalsize. 
The kinds of goals that {\em RTPI} handles are typical of critiquing
systems, systems that provide instructions for performing a task,
etc. These goals may consist of getting the user to perform actions,
refrain from performing actions, use an alternate method to achieve a
goal, or recognize the temporal constraints on actions.



Rules are defined in terms of tree specifications and operators, and
are stylistically similar to the kinds of rules proposed in
\cite{Wanner96}. When all the tree specifications are matched, the
score function of the rule is evaluated. The score function is a
heuristic specific to each rule, and is used to determine which rule
instantiation has the best potential text realization. Scores for
aggregation rules, for example, measure the opportunity to reduce
repetition through aggregation, subsumption, or pronominal reference,
and penalize for paragraph complexity.


Once a rule instantiation is chosen, the system performs any
substitutions, pruning, and moving of branches specified by the rule's
operators. The rules currently in use operate on text plan trees in a
pairwise fashion, and recursively add more text plans to larger,
already integrated plans.


\vspace{-.03in}
\subsection{Classes of Rules}
\vspace{-.03in}

{\em RTPI} has three classes of rules, all of which produce an
integrated text plan from separate text plans. The classes of rules
correlate with the three categories of problems that we identified
from our analysis of TraumaTIQ's critiques, namely, the need to:
1) aggregate communicative goals to achieve more succinct text plans;
2) resolve conflict among text plans; and 3) exploit the relationships
between communicative goals to enhance coherence.

\subsubsection{Aggregation}
\label{sec-agg}

\begin{figure}
\centerline{
\includegraphics[width=3.0in]{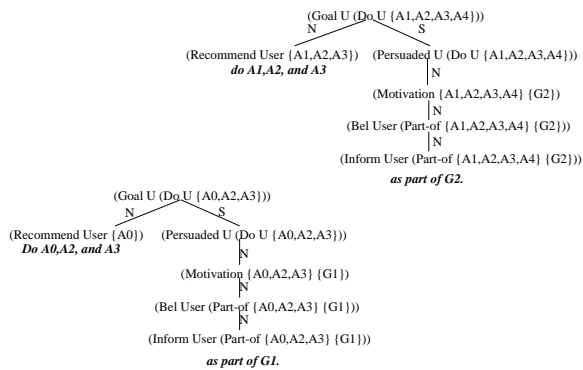}}
\caption{Input to {\em RTPI} (see Figure \ref{agg1}).}
\rule{3.1in}{.010in}
\label{pre-agg1}
\end{figure}

\begin{figure*}[tb]
\centerline{
\includegraphics[width=6.5in]{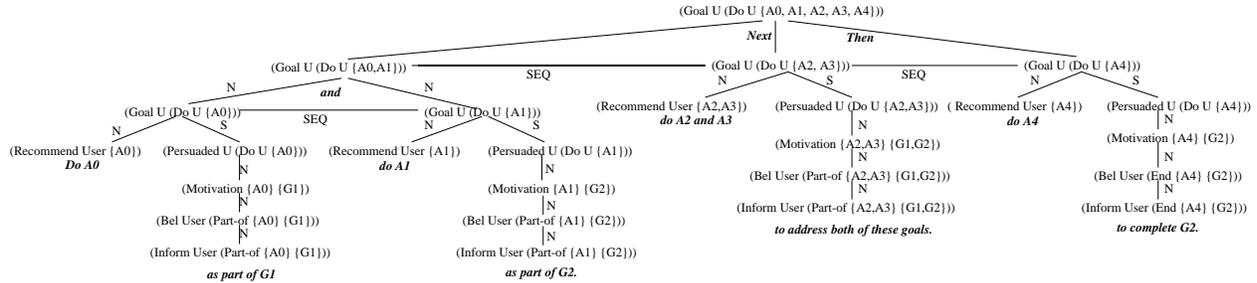}}
\caption{Result of a complex aggregation rule (see Figure \ref{pre-agg1}).}
\rule{6.5in}{.010in}
\label{agg1}
\end{figure*}
\vspace{-.05in}

Our analysis of TraumaTIQ's output showed that one prevalent problem
was informational overlap, i.e. the same actions and objectives often
appeared as part of several different input text plans, and thus the
resulting messages appear repetitious. Aggregation of the
communicative goals associated with these actions and objectives
allows {\em RTPI} to make the message more concise. 


Aggregation of overlapping communicative goals is not usually
straightforward, however, and often requires substantial reorganizing
of the trees.  Our approach was to draw on the
ordered, multi-nuclear \textsc{sequence} relation of RST. We posited
that separate plans with overlapping communicative goals could often
be reorganized as a sequence of communicative goals in a single
plan. The recommended actions can be distributed over the sequentially
related goals as long as the new plan captures the relationships
between the actions and their motivations given in the original plans.

For example, one complex class of aggregation is the integration of
text plans that have overlapping actions or objectives, but also
contain actions and objectives that do not overlap. When those
that overlap can be placed together as part of a valid sequence, a
multi-part message can be generated. 
\emph{RTPI} produces an integrated text plan comprised of sequentially
related segments, with the middle segment conveying the shared actions
and their collected motivations. The other segments convey the actions
that temporally precede or follow the shared actions, and are also
presented with their motivations. For example (Fig. \ref{pre-agg2}),
suppose that one text plan has the goal of getting the user to perform
actions ${A0,A2,}$ and $A3$ to achieve $G1$, while a second text plan
has a goal of getting the user to perform ${A1,A2,A3,}$ and $A4$ to
achieve $G2$. Figure~\ref{agg1} presents the text plan resulting from
the application of this rule. Realization of this text plan in English
produces the message:
\vspace{-.05in}
\begin{center}
\begin{tabular}{p{3.0in}}
{\em Do $A0$ as part of $G1$, and $A1$ as part of $G2$.
Next do $A2$ and $A3$ to address both of these goals.
Then do $A4$ to complete $G2$.}
\end{tabular}
\end{center}
This kind of aggregation is especially appropriate in a domain (such as
trauma care) where the clause re-ordering normally applied to enable
aggregation (e.g. Sentence Planner) is restricted by the partial
ordering of sequenced instructions.

\begin{figure*}[bt]
\centerline{
\includegraphics[width=6.5in]{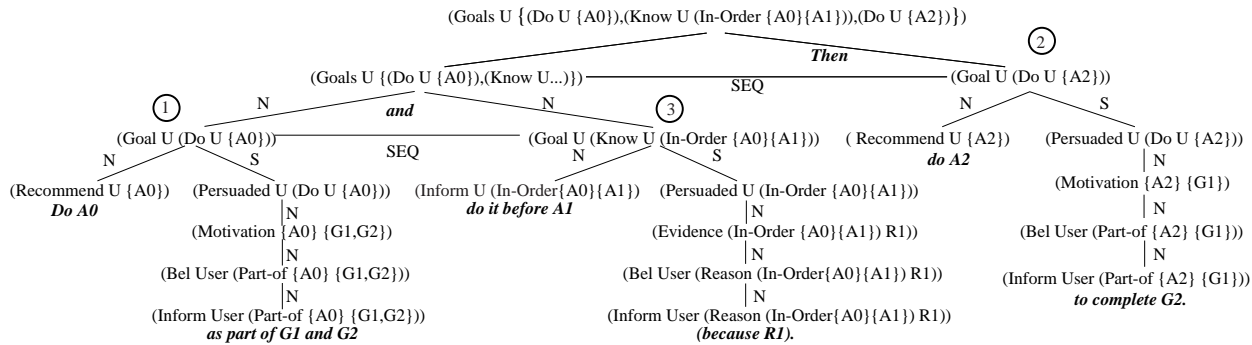}}
\caption{Result of two rules applied to input shown in
Fig. \ref{pre-agg2}.  First, a rule that applies to trees with top level
goals of the form \scriptsize(GOAL USER (DO ...))\normalsize uses two
trees from Fig. \ref{pre-agg2} to make a tree with the two subtrees
labelled (1) and (2). Next, a rule that places scheduling trees (
\scriptsize (GOAL U (KNOW U (IN-ORDER ...))) \normalsize) with related
goals inserts a third subtree (3), in this case the entire scheduling
tree. A domain specific realizer traverses the tree and inserts cue
words and conjunctions based on relations.}
\rule{6.5in}{.010in}
\label{agg2}
\end{figure*}

\begin{figure}[bt]
\centerline{
\includegraphics[width=3.2in]{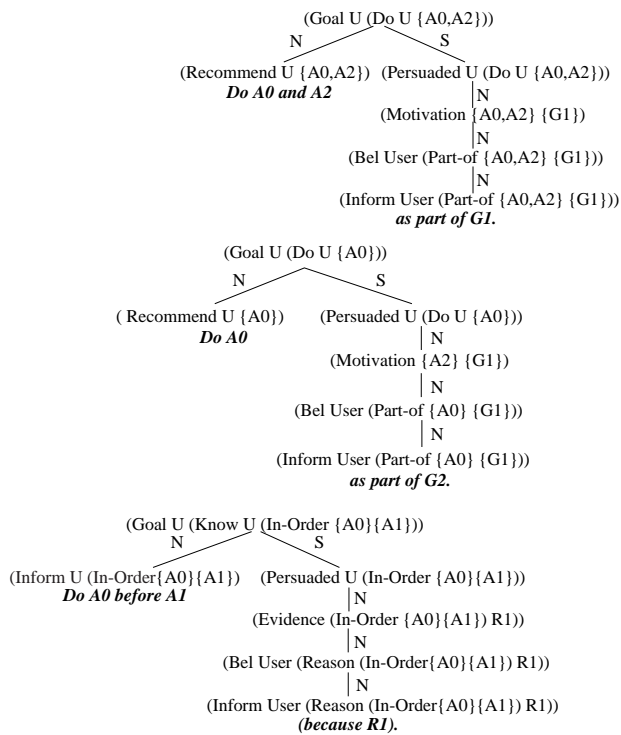}}
\caption{Input to {\em RTPI} (see Figure \ref{agg2}).}
\rule{3.1in}{.010in}
\label{pre-agg2}
\end{figure}

{\em RTPI} can also handle
aggregation when actions or objectives are shared between different
kinds of communicative goals. The bottom part of Figure~\ref{agg-text}
is the text realized from a text plan that was produced by the
application of two rules to three initial text plans: one rule that
applies to trees of the same form, and one that applies to two
distinct forms. The first rule aggregates the communicative goal
\scriptsize\verb+(GOAL USER (DO USER check_med_allergies))+
\normalsize that exists in two of the text plans. The second rule
looks for overlap between the communicative goal of getting the user
to do an action and the goal of having the user recognize a temporal
constraint on actions. The application of these two rules to the text
plans of the three initial messages shown in the top part of
Figure~\ref{agg-text} creates the integrated text plan shown in
Figure~\ref{agg2} whose English realization appears in the bottom part
of Figure~\ref{agg-text}.

{\em RTPI}'s parameter settings capture aspects of the environment in
which the messages will be generated that will affect the kind of
aggregation that is most appropriate. The settings for aggregation
determine whether {\em RTPI} emphasizes actions or objectives. In the
latter case (appropriate in the trauma decision-support environment),
an arbitrary limit of three is placed on the number of sequentially
related segments in a multi-part message, though each segment can
still address multiple goals. This allows the reorganization of
communicative goals to enable aggregation while maintaining focus on
objectives.

 
\subsubsection{Resolving Conflict}
\label{sec-conflict}

The ability to recognize and resolve conflict is required in a text
planner because \emph{both} the appearance and resolution of conflict
can be the result of text structure. {\em RTPI} identifies and resolves
a class of domain-independent conflict, with the resolution strategies
dependent upon the social relationship between the user and the
system. In addition, the system allows the user to add rules for
domain-specific classes of conflict.

One class of conflict that can best be resolved at the text planning
level results from implicit messages in text. Resolving conflict of
this kind within independent modules of a critiquing system would
require sharing extensive knowledge, thereby violating modularity
concepts and making the planning process much more complex.

For example, suppose that the user has conveyed an intention to
achieve a particular objective by performing act $A_{u}$. One system
module might post the communicative goal of getting the user to
recognize that act $A_{p}$ must precede $A_{u}$, while a different
module posts the goal of getting the user to achieve the objective by
executing $A_{s}$ instead of $A_{u}$. While each of these
communicative goals might be well-motivated and coherent in isolation,
together they are incoherent, since the first presumes that $A_{u}$
{\em will} be executed, while the second recommends retracting the
intention to perform $A_{u}$. A text planner with access to both of
these top-level communicative goals {\em and their text plans} can
recognize this implicit conflict and revise and integrate the text
plans to resolve it.

There are many ways to unambiguously resolve this class of implicit
conflict. Strategy selection depends on the \emph{social relationship}
between the system and the user, as captured by three of {\em RTPI}'s
parameter settings. This relationship is defined by the relative
levels of knowledge, expertise, and responsibility of the system and
user. Three strategies used by our system, and their motivations, are:
\begin{description}
\vspace{-.05in}
\item[I.] Discard communicative goals that implicitly conflict with a
system recommendation. In the above example, this would result in a
text plan that only recommends doing $A_{s}$ instead of
$A_{u}$. This strategy would be appropriate if the system is an
expert in the domain, has full knowledge of the current situation, and
is the sole arbiter of correct performance.
\vspace{-.05in}
\item[II.]  Integrate the text plan that implicitly conflicts with the
system recommendation as a concession that the user may choose not to
accept the recommendation. This strategy is appropriate if the 
system is an expert in the domain, but the user has better knowledge
of the current situation and/or retains responsibility for selecting
the best plan of action. Decision support is such an environment. The
top half of Figure~\ref{conflict} presents two TraumaTIQ critiques
that exhibit implicit conflict, while the bottom part presents the
English realization of the integrated text plan, which uses a
\textsc{concession} relation to achieve coherence.
\vspace{-.05in}
\item[III.] Present the system recommendation as an
alternative to the user plan. This may be appropriate if the
parameters indicate the user has more complete knowledge and more
expertise.
\end{description}
\vspace{-.05in}

\subsubsection{Exploiting Related Goals}
\label{sec-exploit}

Occasionally two text plans may exhibit no conflict, yet the
relationships between their communicative goals can be exploited to
produce more coherent text. For example, consider the following two
individual critiques produced by TraumaTIQ:
\vspace{-.05in}
\begin{center}
\begin{tabular}{p{3.0in}}
{\em Caution: do a peritoneal lavage immediately as part of ruling out
abdominal bleeding.} \\ \\
{\em Do not reassess the patient in 6 to 24 hours until after doing
a peritoneal lavage.  The outcome of the latter may affect the need to
do the former.}
\end{tabular}
\end{center}
While the two critiques do not conflict, {\em RTPI}'s rules exploit the
relation between the communicative goals in their respective text
plans to produce a more concise and coherent message. In particular,
one of {\em RTPI}'s rules recognizes the interaction between an initial
plan to get the user to perform an action $A_{s}$, and a second plan that
gets the user to recognize a dependency between $A_{s}$ and another
action. This rule creates a text plan for the message:
\begin{center}
\begin{tabular}{p{3in}}
{\em Do a peritoneal lavage immediately as part of ruling out
abdominal bleeding.  Use the results of the peritoneal lavage
to decide whether to reassess the patient in 6 to 24 hours.}
\end{tabular}
\end{center} 

\vspace{-.03in}
\subsection{Trailing Comments}
\label{sec-trail}
\vspace{-.03in}

Occasionally when several text plans are integrated into a single text
plan, another text plan that overlaps with the integrated plan will
remain outside the new plan because the scoring function for the
applicable rule was too low to allow it to combine. This is typically
because an effort to integrate such a text plan would create a message
so complex that the heuristic deemed it inappropriate.

However, once concepts have been introduced in the integrated text
plan, focusing heuristics \cite{McKeown85} suggest that other text
plans containing these concepts be included in the integrated plan as
well.  Rather than restructure the result of our transformation
(against the advice of our heuristic), we append them to the end of
the message.  Thus we refer to them as {\em trailing comments}.

Unfortunately, when the communicative goal is to get the user to
perform an action, trailing comments that refer to such actions have
the potential to erroneously suggest new instances of actions.  Our
solution to this problem is implemented in the text realization
templates, where we (1) make the focused action the subject of the
sentence, reflecting its {\em given} status in the discourse, (2)
utilize clue words to call attention to its occurrence earlier in the
message and to the new information being conveyed, and (3) subordinate
other concepts presented with the focused concept by placing them in a
phrase introduced by the cue words ``along with''. In one such example
from the trauma domain, the main text plan contains the communicative
goal of getting the user to perform several actions, including a
laparotomy. A \textsc{sequence} relation is used to adjoin an
overlapping text plan as a trailing comment, and this additional
communicative goal is realized in English as (clue words underlined):
\vspace{-.05in}
\begin{center}
\begin{tabular}{p{3in}}
{\em \underline{Moreover}, doing the laparotomy is \underline{also}
indicated, \underline{along with} repairing the left diaphragm, to
treat the lacerated left diaphragm.}
\end{tabular}
\end{center}

\vspace{-.05in}
\section{Algorithm}
\label{sec-alg}
\vspace{-.05in}

\begin{figure}
\begin{center}
\begin{tabular} {p{3.0in}}
\underline{TraumaTIQ critiques:}\\
{\em Performing local visual exploration of all abdominal wounds is
preferred over doing a peritoneal lavage for ruling out a suspicious
abdominal wall injury.} \\ \\{\em Please remember to check for
laparotomy scars before you do a peritoneal lavage.}\\ \\ 
\underline{Message from \emph{RTPI} integrated plan:}\\
{\em Performing local visual exploration of all abdominal wounds is
preferred over doing a peritoneal lavage for ruling out a suspicious
abdominal wall injury. However, if you do a peritoneal lavage, then
remember to first check for laparotomy scars.}
\end{tabular}
\end{center}
\caption{Conflict resolution.}
\rule{3.1in}{.010in}
\label{conflict}
\end{figure}

{\em RTPI} performs rule-based integration of a set of RST-style
trees. Rules are applied in an order designed to maximize derived
benefit. The system first applies the rules that resolve
conflict, since we hypothesize that the presence of conflict will most
seriously hamper assimilation of a message. Next, the rules that
exploit relations between text plans are tried because they enhance
coherence by explicitly connecting different communicative goals. Then
the aggregation rules are applied to improve conciseness. Finally, the
rules for trailing comments reduce the number of disconnected message
units.

The algorithm is both greedy and anytime \cite{garvey93c}; it takes
the best result from a single application of a rule to a set of text
plans, and then attempts to further apply rules to the modified
set. The rule instantiation with the highest heuristic score is chosen
and the rule's operator is applied to the trees using those
bindings. Since the rules are designed to apply incrementally to a
set, every application of a rule results in an improvement in the
conciseness or coherence of the tree set, and the tree set is always a
viable set of text plans. The user can thus set a time limit for
processing of a tree set, and the algorithm can return an improved set
at any time. In practice, however, the processing has never taken more
than 1-2 seconds, even for large (25 plans) input sets.

\vspace{-.05in}
\section{Results}
\vspace{-.05in} 
We tested {\em RTPI} using the corpus of critiques generated by
TraumaTIQ. A set of critiques was extracted from the middle of each of
48 trauma cases, and RST-style text plans were automatically generated
for all the critiques. Then {\em RTPI} ran each set, and messages
resulting from a template-based realization of {\em RTPI}'s text plans
were analyzed for conciseness and coherence. We are currently using
templates for sentence realization since we have been working in the
domain of trauma care, where fast real-time response is essential.

There was a 18\% reduction in the average number of individual text
plans in the 48 sets examined. The results for individual sets ranged
from no integration in cases where all of the text plans were
independent of one another, to a 60\% reduction in sets that were
heavily inter-related. More concise messages also resulted from a 12\%
reduction in the number of references to the diagnostic and
therapeutic actions and objectives that are the subject of this
domain. The new text plans also allowed some references to be replaced
by pronouns during realization, making the messages shorter and more
natural.

To evaluate coherence, messages from twelve cases\footnote{The
evaluation examples consisted of the first eleven instances from the
test set where {\em RTPI} produced new text plans, plus the first
example of conflict in the test set.} were presented, in randomly
ordered blind pairs, to three human subjects not affiliated with our
project. The written instructions given to the subjects instructed
them to note whether one set of messages was more comprehensible, and
if so, to note why. Two subjects preferred the new messages in 11 of
12 cases, and one subject preferred them in all cases. All subjects
{\em strongly} preferred the messages produced from the integrated
text plan 69\% of the time.

\vspace{-.05in}
\section{Summary}
\vspace{-.05in}
Integration of multiple text plans is a task that will become
increasingly necessary as independent modules of sophisticated systems
are required to communicate with a user. This paper has presented our
rule-based system, {\em RTPI}, for accomplishing this task.
{\em RTPI} aggregates communicative goals to achieve more succinct
text plans, resolves conflict among text plans, and exploits the
relations between communicative goals to enhance coherence.


{\em RTPI} successfully integrated multiple text plans to improve
conciseness and coherence in the trauma care domain. We will further
explore the application of {\em RTPI}'s domain-independent rules by
applying the system to a different domain. We would also like to
develop more domain-independent and some domain-dependent rules, and
compare the fundamental characteristics of each.
\vspace{-.10in}
\bibliography{references}
\end{document}